\documentclass[10pt,twocolumn,letterpaper]{article}

\usepackage{cvpr}
\usepackage{times}
\usepackage{epsfig}
\usepackage{graphicx}
\usepackage{amsmath}
\usepackage{amssymb}
\usepackage[numbers,sort&compress]{natbib}
\usepackage{multirow}
\usepackage{gensymb}
\usepackage{caption}
\usepackage{subcaption}
\usepackage[table]{xcolor}
\usepackage{pifont}
\usepackage{arydshln}
\usepackage{makecell}
\usepackage{url}

\iftrue
\newcommand{\davide}[1]{\textcolor{blue}{\bf [DAVIDE: #1]}}
\newcommand{\iris}[1]{\textcolor{green}{\bf [IRIS: #1]}}
\newcommand{\joe}[1]{\textcolor{magenta}{\bf [JOE: #1]}}
\newcommand{\todo}[1]{\textcolor{red}{[#1]}}
\else
\newcommand{\davide}[1]{}
\newcommand{\iris}[1]{}
\newcommand{\joe}[1]{}
\newcommand{\todo}[1]{}
\fi
\usepackage[pagebackref=false,breaklinks=true,letterpaper=true,colorlinks,bookmarks=false]{hyperref}

\cvprfinalcopy 

\def\cvprPaperID{7862} 

\ifcvprfinal\fi
\begin{document}

\title{Combining detection and tracking for human pose estimation in videos}

\author{Manchen Wang, $\;$ Joseph Tighe, $\;$ Davide Modolo\\ 
AWS Rekognition \\
{\tt\small manchenw,tighej,dmodolo@amazon.com}}

\maketitle
\thispagestyle{empty}

\begin{abstract}
We propose a novel top-down approach that tackles the problem of multi-person human pose estimation and tracking in videos. In contrast to existing top-down approaches, our method is not limited by the performance of its person detector and can predict the poses of person instances not localized. It achieves this capability by propagating known person locations forward and backward in time and searching for poses in those regions. Our approach consists of three components: (i) a Clip Tracking Network that performs body joint detection and tracking simultaneously on small video clips; (ii) a Video Tracking Pipeline that merges the fixed-length tracklets produced by the Clip Tracking Network to arbitrary length tracks; and (iii) a Spatial-Temporal Merging procedure that refines the joint locations based on spatial and temporal smoothing terms. Thanks to the precision of our Clip Tracking Network and our merging procedure, our approach produces very accurate joint predictions and can fix common mistakes on hard scenarios like heavily entangled people. Our approach achieves state-of-the-art results on both joint detection and tracking, on both the PoseTrack 2017 and 2018 datasets, and against all top-down and bottom-down approaches.
\end{abstract}

\vspace{-4mm}
\section{Introduction}

Multi-person human pose tracking is the dual-task of detecting the body joints of all the people in all video frames and linking them correctly over time. 
The ability to detect body joints has improved considerably in the last several years~\cite{andriluka14cvpr,pishchulin16cvpr,wei16cvpr,newell16eccv,insafutdinov16eccv,newell2017associative,cao2018openpose,chen2018cascaded, he2017mask, sun2019deep} thanks in part to the availability of large scale public image datasets like MPII~\cite{andriluka14cvpr} and MS COCO~\cite{lin2014microsoft}. 
These approaches can be mostly classified into two categories, depending on how they operate: bottom-up approaches~\cite{andriluka14cvpr, wei16cvpr, cao2018openpose, insafutdinov16eccv, newell2017associative, pishchulin16cvpr} first detect individual body joints and then group them into people; while top-down approaches~\cite{chen2018cascaded, he2017mask, sun2019deep} first detect every person in an image and then predict each person's body joints within their bounding box location. 

\begin{figure}
        \includegraphics[width=\linewidth]{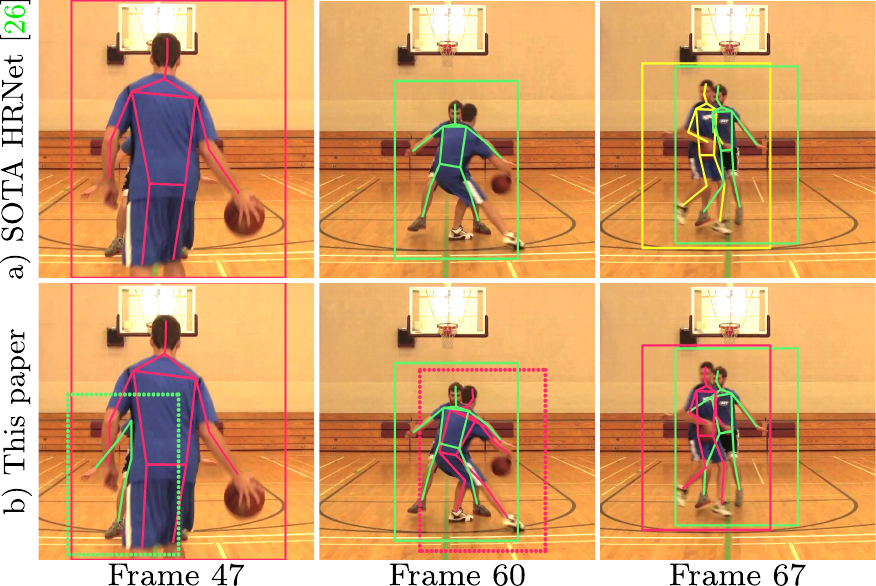}
        \vspace{-5mm}
    \caption{\small \it Top-down approaches like HRNet rely heavily on the performance of their person detector, which sometimes fails on highly occluded people (frames 47, 60), and occasionally make mistakes on highly entangled people (frame 67). Our approach overcomes these limitations by propagating bounding boxes over time (drawn with dotted lines) and by predicting multiple pose hypothesis for each person and smartly selecting the best one. \vspace{-5mm}}
    \label{fig:teaser}
\end{figure}

Largely thanks to advancements in object class detection~\cite{he2017mask,dai17iccv,singh18nips}, top-down approaches~\cite{sun2019deep} have achieved better pose estimation performance on images than bottom-up methods. By taking advantage of robust person detectors, these approaches can focus on the task of joint detection within bounding box regions, and not have to deal with large scale variations and the problem of grouping joints into people that bottom-up methods do. Despite these positive results on image datasets, top-down methods do not perform as well on videos and were recently outperformed by a bottom-up approach~\cite{raaj2019efficient}. 
We attribute this to the fact that detecting people bounding boxes in videos is a much harder task than on images. While images often capture people ``posing'', videos inherently contain atypical types of occlusion, viewpoints, motion blur and poses that make object detectors occasionally fail (e.g., in fig.~\ref{fig:teaser}{\color{red}a}, the detector is not able to localize the highly occluded person instances in the first two frames).

We propose a novel top-down approach that overcomes these problems and enables us to reap the benefits of top down methods for multi-person pose estimation in videos.
We detect person bounding boxes on each frame and then propagate these to their neighbours. Our intuition is that if a person is present at a specific location in a frame, they should still be at approximately that location in the neighbouring frames, even when the detector fails to find them.
In detail, given a localized person bounding box, we crop a spatial-temporal tube from the video centered at that frame and location. Then, we feed this tube to a novel {\it Clip Tracking Network} that estimates the locations of all the body joints of that person in all the frames of the tube. To solve this task, our Clip Tracking Network performs body joint detection and tracking simultaneously. This has two benefits: (i) by solving these tasks jointly, our network can better deal with unique poses and occlusions, and (ii) it can compensate for missed detections by predicting joints in all frames of the spatial-temporal tube, even for frames where the person was not detected. To construct this Clip Tracking Network, we extend the state-of-the-art High-Resolution Network (HRNet)~\cite{sun2019deep} architecture to the task of tracking, using 3D convolutions that are carefully designed to help learn the temporal correspondence between joints. 

The Clip Tracking Network operates on fixed length video clips and produces multi-person pose tracklets. We combine these tracklets into pose tracks for arbitrary length videos in our {\it Video Tracking pipeline}, by first generating temporally overlapping tracklets and then associating and merging the pose detections in frames where the tracklets overlap. When merging tracklets into tracks, we use the multiple pose detections in each frame in a novel consensus-based {\it Spatio-temporal merging} procedure to estimate the optimal location of each joint.
This procedure favours hypotheses that are spatially close to each other and that are temporally smooth. This combination is able to correct mistakes on highly entangled people, leading to more accurate predictions, as in frame~67 of fig.~\ref{fig:teaser}{\color{red}}: while ~\cite{sun2019deep} wrongly selects the yellow player's left knee as the prediction for the green player's right knee (\ref{fig:teaser}{\color{red}a}), our procedure is able to correct this mistake and predict the correct location (\ref{fig:teaser}{\color{red}b}).

When compared to the literature, our approach achieves state-of-the-art results for both body joint detection and tracking, on the PoseTrack 2017 and 2018 video datasets~\cite{PoseTrack}, not only against top-down approaches, but also against bottom-up ones. The improvement is consistent and often significant; for example, error on body joint detection reduces by 28\% PoseTrack 2017 and error on body joint tracking by 9\% on PoseTrack 2018. Furthermore, we also present an extensive ablation study of our approach, where we validate its components and our hyperparameter choices. 

The rest of the paper is organized as follows: in sec.~\ref{sec:rl} we present our related work; then, in sec.~\ref{sec:method} we present our three contributions: (i) our novel clip tracking network (sec.~\ref{sec:3DHRnet}), (ii) our tracking pipeline (sec.~\ref{sec:tracking}) and (iii) our spatial-temporal merging procedure (sec.~\ref{sec:refinement}). Finally, we present our experiments in sec.~\ref{sec:exp} and conclude in sec.~\ref{sec:concl}.

\section{Related Work}\label{sec:rl}
\subsection{Human pose estimation in images}
Recent human pose estimation methods can be classified into bottom-up and top-down approaches, depending on how they operate. 
{\it Bottom-up approaches} \cite{cao2018openpose, insafutdinov16eccv, newell2017associative, pishchulin16cvpr} first detect individual body joints and then group them into people. 
On the other hand, {\it top-down approaches}~\cite{chen2018cascaded,papandreou17cvpr, he2017mask, sun2019deep}, first detect people bounding boxes and then predict their joint locations within each region. 
Top-down approaches have the advantage of not needing any joint grouping and because the input images they operate on are crops from detectors, they do not have to be robust to large scale variations. However, top-down approaches suffer from the limitations of the person detector: when it fails (i.e., a person is not localized), the joints on that person cannot be recovered. Bottom-up approaches do not have this reliance on a detector and they can predict any joint; however they suffer from the difficult tasks of joint detection across large scale variations and joints grouping. In this work we try to take the best of both words and propose a novel top-down approach for videos that recovers from the detector's misses by exploring and propagating information temporally.

We build upon the HRNet of Sun et al.~\cite{sun2019deep}.
This was originally proposed for human pose estimation, achieving state-of-the-art results in images. Recently, it was then modified to achieve state-of-the-art results on other vision tasks, like object detection~\cite{sun2019arxiv} and semantic segmentation~\cite{wang19arxiv}. In this paper we show how to extend HRNet to human pose estimation and tracking in videos.

\subsection{Human pose estimation and tracking in videos}
Given the image approaches just introduced, it is natural to extend them to multi-person pose tracking in videos by running them on each frame independently and then linking these predictions over time. 
Along these lines, {\it bottom-up methods}~\cite{raaj2019efficient,jin2019multi} build spatial-temporal graphs between the detected joints. Raaj et al.~\cite{raaj2019efficient} did so by extending the spatial Affinity Field image work of Cao et al.~\cite{cao2018openpose} to Spatio-Temporal Affinity Fields (STAF), while Jin et al.~\cite{jin2019multi} extended the spatial Associative Embedding image work of Newell et al.~\cite{newell2017associative} to Spatio-Temporal Embedding. 

On the other hand, {\it top-down methods}~\cite{xiao2018simple,girdhar2018detecttrack} build temporal graphs between person bounding boxes, which are usually simpler to solve.     
SimpleBaseline~\cite{xiao2018simple} first run a person detector on each frame independently and then linked its detections in a graph, where the temporal similarity was defined using expensive optical flow. Detect-and-Track~\cite{girdhar2018detecttrack} instead used a 3D Mask R-CNN approach to detect the joints of a person in a short video clip and then used a lightweight tracker to link consecutive clips together by comparing the location of the detected bounding boxes. 
Like \cite{girdhar2018detecttrack}, our approach also runs inference on short clips in a single forward pass, but it brings many advantages over it: (i) as most top-down approaches, \cite{girdhar2018detecttrack} is limited by its detector's accuracy and it cannot recover from its misses; instead, we propose to propagate detected bounding boxes to neighbouring frames and look for missed people in those regions. (ii) \cite{girdhar2018detecttrack} runs on non-overlapping clips and performs tracking based on person bounding boxes only; instead, we run on overlapping clips and use multiple joint hypothesis in a novel tracking system, that leads to more accurate predictions. (iii) \cite{girdhar2018detecttrack} employs fully 3D convolutional networks, while we show that 3D filters on only part of a network is already sufficient to teach the network to track.

\begin{figure*}
    \centering
       \includegraphics[width=1.0\textwidth]{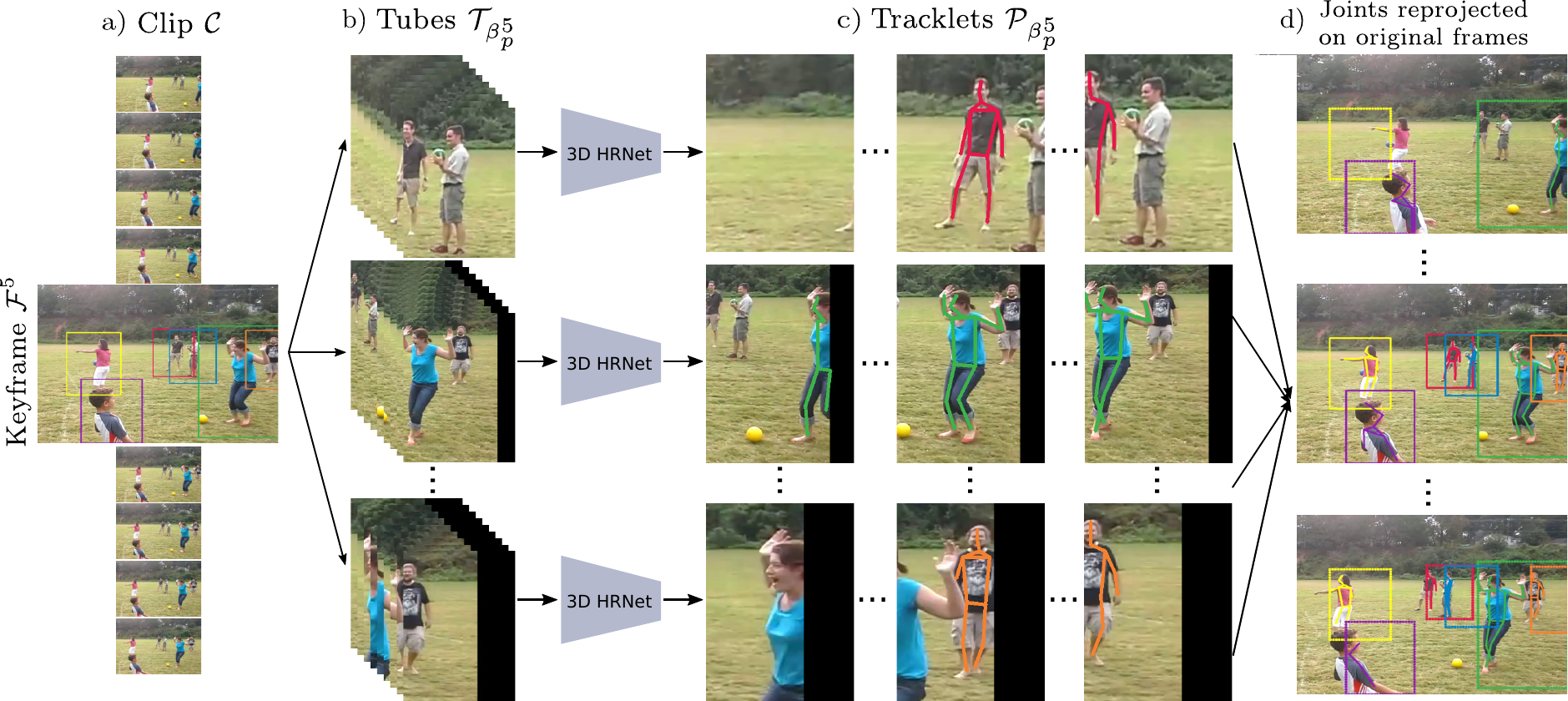}
       \vspace{-4mm}
    \caption{\small \it {\bf Clip Tracking Network}. First, (a) our approach runs a person detector on the keyframe of a short video clip. Then, (b) for each detected person it creates a tube by cropping the region within his/her bounding box from all the frames in the clip. Next, (c) each tube is independently fed into our Clip Tracking Network (3D HRNet), which outputs pose estimates for the same person (the one originally detected in the keyframe) in all the frames of the tube. Finally, (d) we reproject the predicted poses on the original images to show how the model can correctly predict poses in all the frames of the clip, by only detecting people in the keyframe. \vspace{-3mm}}
    \label{fig:clip_tracking_network_fig} 
\end{figure*}
\section{Methodology}\label{sec:method}

At a high level, our method works by first detecting all candidate persons in the center frame of each video clip (i.e. the {\it keyframe}) and then estimating their poses forward and backward in time. Then, it merges poses from different clips in time and space, producing any arbitrary length tracks.
More in details, our approach consist of three major components: {\it Cut, Sew} and {\it Polish}. 
Given a video, we first cut it into overlapping {\it clips} and then run a person detector on their keyframes. 
For each person bounding box detected in a keyframe, a spatial-temporal tube is {\it cut} out at the bounding box location over the corresponding clip.
Given this tube as input, our {\bf Clip Tracking Network} both estimates the pose of the central person in the keyframe, and tracks his pose across the whole video clip (sec.~\ref{sec:3DHRnet}, fig.~\ref{fig:clip_tracking_network_fig}).
We call these {\it tracklets}. 
Next, our {\bf Video Tracking Pipeline} works as a tailor to {\it sew} these tracklets together based on poses in overlapping frames (sec.~\ref{sec:tracking}, fig.~\ref{fig:long_video_tracking_fig}). We call these multiple poses for the same person in same frame {\it hypotheses}.
Finally, {\bf Spatial-Temporal merging} {\it polishes} these predictions using these hypotheses in an optimization algorithm that selects the more spatially and temporally consistent location for each joint  (sec.~\ref{sec:refinement}, fig.~\ref{fig:refinement}).
In the next three sections we present these three components in details.

\subsection{Clip Tracking Network}\label{sec:3DHRnet}
Our Clip Tracking Network performs both pose estimation and tracking simultaneously, on a short video clip. Its architecture builds upon the successful HRNet architecture of Sun et al.~\cite{sun2019deep}. In the next paragraph we summarize the original HRNet design and in the following one we explain how to extend it to tracking.

\paragraph{HRNet for human pose estimation in images.} Given an image, this top-down approach runs a person detector on it, which outputs a list of axis-aligned bounding boxes, one for each localized person. Each of these boxes is independently cropped and fed into HRNet, which consists of four stages of four parallel subnetworks trained to localize all body joints of only the central person in the crop.

The output of HRNet is a set of heatmaps, one for each body joint. Each pixel of these heatmaps indicates the likelihood of ``containing'' a joint. 
As other approaches in the literature~\cite{cao2018openpose, insafutdinov16eccv, newell2017associative, pishchulin16cvpr,chen2018cascaded, he2017mask}, the network is trained using a mean squared error loss function, between the predicted heatmap  $H^{pred}$ and the ground-truth heatmap  $H^{gt}$:
\begin{equation}
L = \frac{1}{K W H}\sum_k^{K} \sum_i^W \sum_j^H \left\lVert H_{ijk}^{pred}-H_{ijk}^{gt} \right\rVert _2^2,
\label{eq:hrnet}
\end{equation}
where $K$ is the number of body joints (keypoints) and $i,j$ the pixel coordinates.  $H^{gt}$ are generated by convolving a 2D Gaussian filter on the annotated location of each joint. 

\paragraph{3D HRNet for video pose estimation and tracking.} Our approach operates on short video clips: {\small $\mathcal{C} = \{\mathcal{F}^{t-\delta}, ..., \mathcal{F}^t, ..., \mathcal{F}^{t+\delta}\}$}. 
First, it runs a person detector on the center frame $\mathcal{F}^t$ and obtains a list of person bounding boxes {\small $\mathcal{B}^t = \{\beta_1^t, ..., \beta_n^t\}$} (fig.~\ref{fig:clip_tracking_network_fig}{\color{red}a}).
Then, for each bounding box {\small $\beta_p^t$}, it creates a tube {\small $\mathcal{T}_{\beta_p^t}$}  by cropping the box region from all frames in the clip {\small $\mathcal{C}$: $\mathcal{T}_{\beta_p^t}= \{ \mathcal{F}_{\beta_p^t}^{t-\delta}, ..., \mathcal{F}_{\beta_p^t}^t, ..., \mathcal{F}_{\beta_p^t}^{t+\delta}\}$} (fig.~\ref{fig:clip_tracking_network_fig}{\color{red}b}). 
Next, it feeds this tube to our video HRNet, which outputs a {\it tracklet} containing all the poses of person $p$ in all the frames of the tube: {\small $\mathcal{P}_{\beta_p^t} = \{\rho_{\beta_p^t}^{t-\delta}, ..., \rho_{\beta_p^t}^t, ..., \rho_{\beta_p^t}^{t+\delta}\}$} (fig.~\ref{fig:clip_tracking_network_fig}{\color{red}c}).
Importantly, all the poses in $\mathcal{P}_{\beta_p^t}$ need to belong to the same person, even when this becomes occluded or moves out of the tube frame (in which case the network should not output any prediction, even if other people are present). This is a difficult task, which requires the network to both learn to predict the location of the joints of the pose and track them through time.

In order to help the network tackle this challenge, we do two things: (i) to account for fast moving people, we enlarge each bounding box by 25\% along both dimensions prior to creating a tube; and (ii) to allow the network to associate people between frames, we inflate the 2D convolutions in the first two stages of HRNet to 3D to help the network learn to track. Specifically, in the first stage 
we use {\small 3$\times$1$\times$1, 1$\times$3$\times$3 and 1$\times$1$\times$1} filters, while in the second stage we use {\small3$\times$3$\times$3} filters. After this second stage the network has a receptive field that is temporally large enough to observe the whole tube, learn the person's appearance and his/her movements within it. 
Note how our method is similar in spirit to what Jin et al.~\cite{jin2019multi} proposed with their temporal associative embedding, but it is learnt automatically by the network without the need of additional constraints. 
Finally, we train our video HRNet with the same mean squared loss of eq.~\ref{eq:hrnet}, but now computed over all the frames in the clip $\mathcal{C}$:
\begin{equation}
L = \frac{1}{|\mathcal{C}| K W H}\sum_f^{|\mathcal{C}|} \sum_k^{K} \sum_i^W \sum_j^H \left\lVert H_{ijkf}^{pred} - H_{ijkf}^{gt} \right\rVert _2^2
\end{equation}

\subsection{Video Tracking Pipeline}\label{sec:tracking}

\begin{figure}
   \includegraphics[width=\columnwidth]{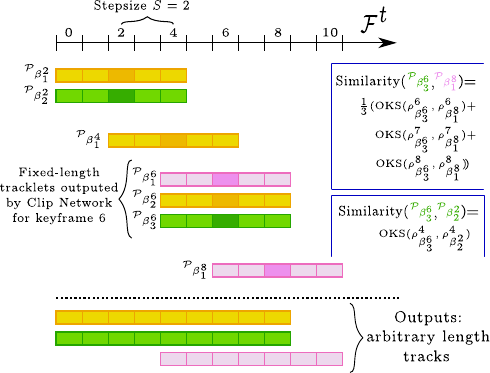}
   \vspace{-7mm}
   \caption{\small \it {\bf Video Tracking Pipeline} merges fixed-length tracklets into arbitrary length tracks by comparing the similarity of their detected poses in the frames the tracklets overlap on.\vspace{-4mm}}
   \label{fig:long_video_tracking_fig} 
\end{figure}

Our Clip Tracking Network outputs a tracklet  $\mathcal{P}_{\beta_p^t}$ for each person $p$ localized at $\beta_p$. However, $p$ may exist beyond the length of  $\mathcal{P}_{\beta_p^t}$ and the duty of our Video Tracking pipeline is to merge tracklets that belong to the same person, thus enabling pose estimation and tracking on any arbitrary length video (fig.~\ref{fig:long_video_tracking_fig}). 
Our pipeline merges two fixed-length tracklets if their predicted poses on overlapping frames are similar (e.g., in fig.~\ref{fig:long_video_tracking_fig}, $\mathcal{P}_{\beta_1^2}$ and $\mathcal{P}_{\beta_1^4}$ overlap on frames 2-4).

We generate these overlapping tracklets by running our Clip Tracking Network on clips of length $|\mathcal{C}|$ from keyframes sampled every $S$ ({\it stepsize})  frames with $S<|\mathcal{C}|$. 

We model the problem of merging tracklets that belong to the same person as a bipartite graph based energy minimization problem, which we solve using the Hungarian algorithm~\cite{kuhn1955hungarian}.
As a similarity function between two overlapping tracklets, we compute Object Keypoint Similarity (OKS)~\cite{lin2014microsoft,ronchi17iccv} between their poses (reprojected on the original coordinate space, fig.~\ref{fig:clip_tracking_network_fig}{\color{red}d}) on their overlapping frames. 
For example, in fig.~\ref{fig:long_video_tracking_fig} tracklets {\small $\mathcal{P}_{\beta_3^{6}}$ and $\mathcal{P}_{\beta_1^{8}}$} are computed on tubes generated from keyframes 6 and 10 respectively and of length $|\mathcal{C}|=5$. Under these settings, these tracklets both predict poses for frames 6, 7 and 8 and their similarity is computed as the average OKS over these three frames. On the other hand, tracklets {\small $\mathcal{P}_{\beta_3^{6}}$ and $\mathcal{P}_{\beta_2^{2}}$} only overlap on frame 4 and as such their similarity is computed as the OKS on that single frame. Finally, we take the negative value of this OKS similarity for our minimization problem. 

Note how this formulation is able to overcome the limitation that top-down approaches usually suffer from: missed bounding box detections. Thanks to our propagation of person detections from keyframes to their neighbouring frames (fig.~\ref{fig:clip_tracking_network_fig}{\color{red}b}), we are able to obtain joints predictons even for those frames with missed detections. For example, in fig.~\ref{fig:long_video_tracking_fig} the person detector failed to localize the green person in keyframe 4, but by propagating the detections from keyframes 2 and 6 we are able to obtain a pose estimate for frame 4 as well. In addition, we are also able to link these correctly, thanks to the overlap between these two tracklets. 

\subsection{Spatial-Temporal merging of pose hypotheses}\label{sec:refinement}

Our video tracking pipeline merges tracklets, but it does not deal with merging human poses. For example, in fig.~\ref{fig:long_video_tracking_fig} the approach correctly links all the yellow tracklets, but it does not address the question of what to do with the multiple pose estimates for frame 4 (i.e., $\rho_{\beta_1^2}^4, \rho_{\beta_1^4}^4$ and $\rho_{\beta_2^6}^4$). In this section we present our solution to this problem.

\begin{figure}
 	\begin{center}
    	\includegraphics[width=1.02\columnwidth]{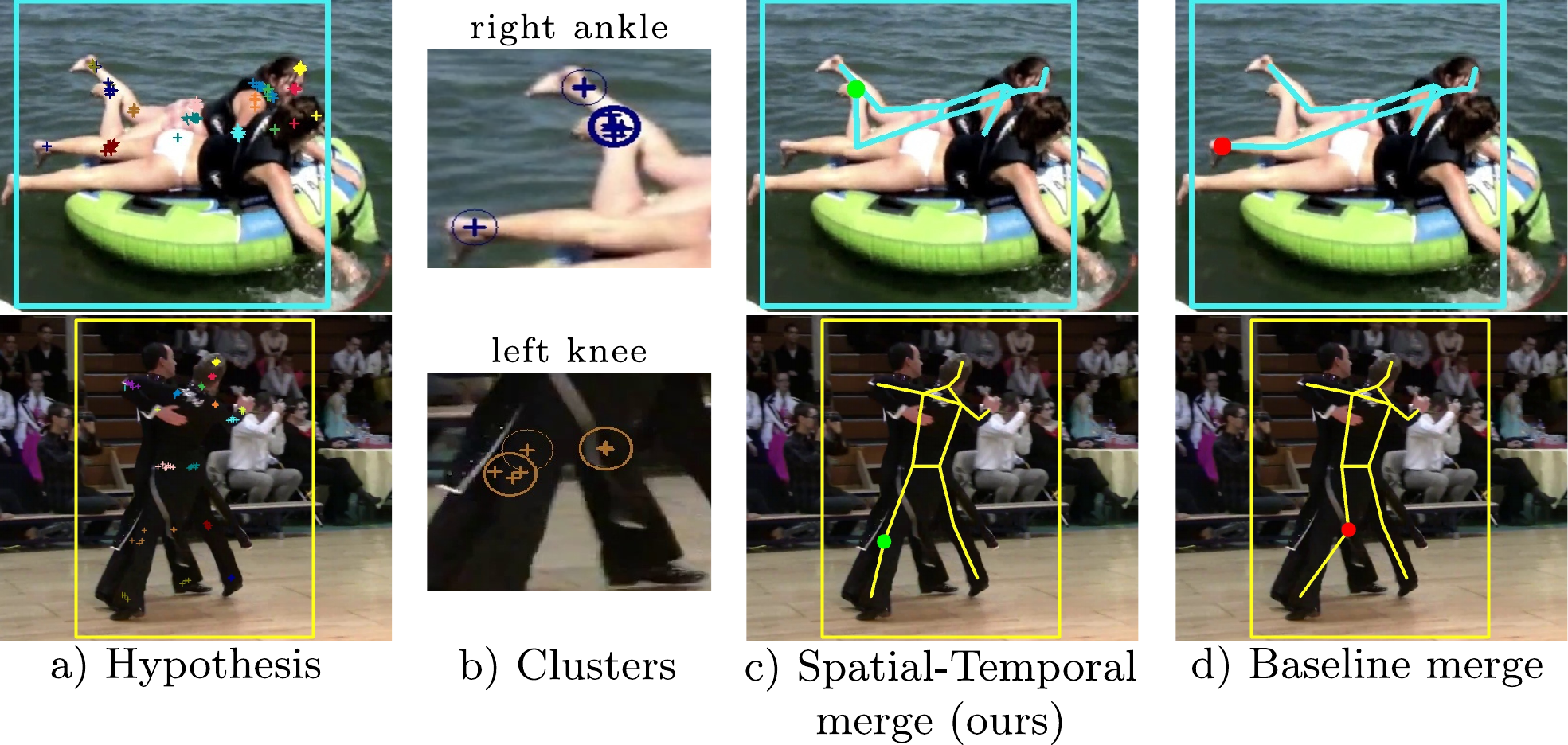}
	\end{center}
	\vspace{-7mm}
    	\caption{\small \it {\bf Merging pose hypotheses.} Our video tracking pipeline runs our Clip Tracking Network on multiple overlapping frames, producing multiple hypotheses for every joint of a person (a). We cluster these hypotheses (b) and solve a spatial-temporal optimization problem on these clusters to estimate the best location of each joint (c). This achieves better predictions than a simple baseline that always pick the hypothesis with the highest confidence score (d), especially on frames with highly entangled people.\vspace{-4mm}}
    \label{fig:refinement}
\end{figure}

Given a set of merged, overlapping tracklets for person $p$, we define {\small $\mathcal{H}_p^t=\{\rho_{\beta_p^{t-\delta}}^{t}, ..., \rho_{\beta_p^{t}}^{t}, ..., \rho_{\beta_p^{t+\delta}}^{t}\}$}, as the {\it pose hypotheses} of $p$ at time $t$. $\mathcal{H}_p^t$ represents the collection of poses for person $p$, generated by our Clip Tracking Network at time $t$ by running on tube crops centered on different keyframes. 
The most straightforward procedure to obtain a single final pose for each person is to simply select, for each joint, the hypothesis $\mathcal{H}_p^t$ with the highest confidence score. We call this {\it Baseline Merge} and, as we show later in our experiments, it achieves competitive performance, already highlighting the power of our Clip Tracking Network. Nevertheless, this procedure occasionally predicts the wrong location when the person of interest is entangled with or occluded by another person, as show in fig.~\ref{fig:refinement}{\color{red}d}. 

To overcome these limitations, we propose a novel method to merge these hypotheses 
(fig.~\ref{fig:refinement}{\color{red}b-c}). Our intuition is that the optimal location for a joint should be the one that is both consistent across the multiple candidates within a frame (spatial constraint) and consistent over consecutive frames (temporal constraint). 
We model the problem of predicting the optimal location for each joint in each frame as a shortest path problem and we solve it using the Dijkstra's algorithm~\cite{dijkstra1959note}. Instead of considering each joint detection as a node in the graph, we operate on clusters obtained by running a mean shift algorithm over joint hypotheses~\cite{comaniciu2002mean}. 
This clusters robustly smooth out noise in the individual hypotheses, while also reducing the graph size leading to faster optimization.
As a similarity function $\phi$ between clusters $c^t$ and $c^{t+1}$ in consecutive frames, we compute a spatial-temporal weighting function that follows the aforementioned intuition: it favours clusters with more hypotheses  and those that have smoother motion across time. 

Formally,

\vspace{-4mm}
\begin{equation}
\vspace{-4mm}
\resizebox{0.9\columnwidth}{!}{
	$\phi(c^t, c^{t+1}) = \underbrace{(|\mathcal{H}| - |c^t|) + (|\mathcal{H}| - |c^{t+1}|)}_\textit{Spatial} + \lambda\underbrace{\left\lVert \mu(c^{t}) - \mu(c^{t+1}) \right\rVert _2^2}_\textit{Temporal},$
	\vspace{-3mm}
}\end{equation}

\noindent where $\mu(c^t)$, $\mu(c^{t+1})$ are the locations of the centers of the clusters, $|c^t|$, $|c^{t+1}|$ their magnitude and $|\mathcal{H}|$ the number of hypotheses. Finally, we balance these spatial and temporal constraints using $\lambda$. 
 
\section{Experiments}\label{sec:exp}

\subsection {Datasets and Evaluation}
We experiment with PoseTrack~\cite{PoseTrack}, which is a large-scale benchmark for human pose estimation and tracking in video. It contains challenging sequences of highly articulated people in dense crowds performing a wide range of activities.
We experiment on both the 2017 and 2018 versions of this benchmark. 
{\bf PoseTrack2017} contains 250 videos for training, 50 for validation and 214 for test. {\bf PoseTrack2018} further increased the number of videos of the 2017 version to a total of 593 for training, 170 for validation and 375 for test. 
These datasets are annotated with 15 body joints, each one defined as a point and associated to a unique person id. Training videos are annotated with a single dense sequence of 30 frames, while validation videos also provide annotations for every forth frame, to enable the evaluation of longer range tracking. 

We evaluate our models using the standard human pose estimation~\cite{pishchulin16cvpr,lin2014microsoft,ronchi17iccv} and tracking~\cite{milan2016mot16,PoseTrack} metrics: joint detection performance is expressed in terms of {\bf average precision (AP)}, while tracking performance in terms of {\bf multi object tracking accuracy (MOTA)}. We compute these metrics independently on each body joint and then obtain our final performance by averaging over the joints. As done in the literature~\cite{girdhar2018detecttrack, xiao2018simple, sun2019deep}, when we evaluate on the validation sets of these datasets, we compute AP on all the localized body joints, but we threshold low confidence predictions prior to computing MOTA.
For our experiments we learn a per-joint threshold on a hold out set of the training set. Moreover, we remove very short tracklets ($<5$ frames) and tiny bounding boxes ($W*H<3200$), as these often capture not annotated, small people in the background.

\subsection {Implementation details}\label{sec:impl_details}

\noindent{\bf 3D Video HRNet.} 
Prior to inflating a 2D HRNet to our 3D version, we pre-train it for image pose estimation on the PoseTrack dataset (2017 or 2018, depending on what set we evaluate the models on). This step enables the network to learn the task of localizing body joints, so that during training on videos it can focus on learning to track. We inflate the first two stages of HRNet using ``mean'' initialization~\cite{carreira2017quo, feichtenhofer2016spatiotemporal, girdhar2018detecttrack}, which replicates the 2D filters and normalizes them accordingly. We use stepsize $S=1$, as it produces the highest number of pose hypotheses, and clips of $|\mathcal{C}|=9$ frames, so that the model can benefit from important temporal information. 
We use the same hyperameters of ~\cite{sun2019deep}, but we train 3D HRNet for 20 epochs and decrease the learning rate two times after 10 and 15 epochs, respectively (1e-4 $\rightarrow$ 1e-5 $\rightarrow$ 1e-6). 
Finally, during inference we follow the procedure of \cite{sun2019deep, xiao2018simple}: we run on both the original and the flipped image and average their heatmaps. 

\noindent{\bf Person detector.} We use a ResNet-101 SNIPER~\cite{singh18nips} detector to localize all the person instances. We train it on the MS COCO 2017 dataset~\cite{lin2014microsoft} and achieve an AP of 57.9 on the ``person'' class on COCO {\it minival}, which is similar to that of other top-down approaches~\cite{xiao2018simple,yu2018multi}.

\noindent{\bf Merging pose hypotheses.} We follow the PoseTrack evaluation procedure to determine a good size estimate for our clusters. This procedure considers a prediction correctly, if the $L_2$ distance between that prediction and the closest ground truth is within a radius defined as 50\% of the head size of the person. We use the same radius for our clusters. Moreover, we set $\lambda=0.1$ to give equal importance to the spatial and temporal components, as the latter has approximately $10\times$ the magnitude of the former.

\subsection{Comparisons with the state-of-the-art} \label{sec:sota}
We compare our approach with the state-of-the-art (SOTA) methods in the literature on body joints detection and tracking, on the validation sets of PoseTrack2017 (tables~\ref{tab:sota_map_posetrack2017} and \ref{tab:sota_mota_posetrack2017}) and PoseTrack2018 (tables~\ref{tab:sota_map_posetrack2018} and \ref{tab:sota_mota_posetrack2018}).
Our approach achieves SOTA results on both metrics, on both datasets and against both top-down and bottom-up approaches. In some cases, the improvement over the SOTA is substantial: +6.5 mAP on PoseTrack2017 (which corresponds to 28\% in error reduction), and +3.0 MOTA on PoseTrack2018 (9\% in error reduction). When compared to only top-down approaches, which is the category this approach belongs to, the improvement in MOTA is even more significant, up to +6.2 on PoseTrack2017 (18\% in error reduction) over the winner of the last PoseTrack challenge (FlowTrack, 65.4 vs 71.6), showing the importance of performing joint detection and tracking simultaneously. 

Next, we evaluate our approach on the test sets of PoseTrack 2017 (table~\ref{tab:sota_posetrack2017_test}) and PoseTrack 2018 (table~\ref{tab:sota_posetrack2018_test}). The annotations for these sets are private and we obtained our results by submitting our predictions to the evaluation server~\cite{posetrack2017}. 
Again, our approach achieves the best tracking results on both test sets (+3 MOTA) and on par to SOTA results on joint detection, even though our model is actually trained on less data than the competitors on PoseTrack2018. 

\begin{table}
\begin{center}
\resizebox{\columnwidth}{!}{
\begin{tabular}{|c|l|c|c|c|c|c|c|c|c|}
\cline{2-10}
\multicolumn{1}{c|}{} & Method & Head & Sho & Elb & Wri & Hip & Kne & Ank & \bf  Avg \\
\hline
\multirow{4}{*}{\rotatebox{90}{Bottom-up}} 
& JointFlow~\cite{doering2018joint}									& - & - & - & - & - & - & - & 69.3\\
& TML++~\cite{hwang2019pose}                                          	           & - & - & - & - & - & - & - & 71.5\\
& STAF~\cite{raaj2019efficient}                                          	           & - & - & - & 65.0 & - & - & 62.7 & 72.6\\ 
& STEmbedding~\cite{jin2019multi}                                          	           & 83.8 & 81.6 & 77.1 & 70.0 & 77.4 & 74.5 & 70.8 & 77.0\\ \cline{2-10}
\hline
\multirow{6}{*}{\rotatebox{90}{Top-down}} 
& Detect\&Track~\cite{girdhar2018detecttrack}           & 67.5 & 70.2 & 62.0 & 51.7 & 60.7 & 58.7 & 49.8 & 60.6 \\ 
& PoseFlow~\cite{xiu2018poseflow}          						  & 66.7 & 73.3 & 68.3 & 61.1 & 67.5 & 67.0 & 61.3 & 66.5 \\
& FastPose~\cite{zhang2019fastpose}                                                & 80.0 & 80.3 & 69.5 & 59.1 & 71.4 & 67.5 & 59.4 & 70.3 \\ 
& FlowTrack~\cite{xiao2018simple}                                           	           & 81.7 & 83.4 & 80.0 & 72.4 & 75.3 & 74.8 & 67.1 & 76.7 \\ 
& HRNet~\cite{sun2019deep}                                          	           & 82.1 & 83.6 & 80.4 & 73.3 & 75.5 & 75.3 & 68.5 & 77.3 \\
& \cellcolor[gray]{0.9}Our approach                      & \cellcolor[gray]{0.9} \bf 89.4 & \cellcolor[gray]{0.9}\bf 89.7 & \cellcolor[gray]{0.9}\bf 85.5 & \cellcolor[gray]{0.9}\bf 79.5 & \cellcolor[gray]{0.9}\bf 82.4 & \cellcolor[gray]{0.9}\bf 80.8 & \cellcolor[gray]{0.9}\bf 76.4 & \cellcolor[gray]{0.9}\bf 83.8\\
\hline
\end{tabular}}
\end{center}
\vspace{-7mm}
\caption{\small \it Joint detection ({\bf AP}) on {\bf PoseTrack2017 val}. \vspace{-2mm}}
\label{tab:sota_map_posetrack2017}
\end{table}

\begin{table}
\begin{center}
\resizebox{\columnwidth}{!}{
\begin{tabular}{|c|l|c|c|c|c|c|c|c|c|}
\cline{2-10}
\multicolumn{1}{c|}{} & Method & Head & Sho & Elb & Wri & Hip & Kne & Ank & \bf Avg\\
\hline
\multirow{4}{*}{\rotatebox{90}{Bottom-up}} 
& JointFlow~\cite{doering2018joint}						& - & - & - & - & - & - & - & 59.8\\
& TML++~\cite{hwang2019pose}                                          	           & 75.5 & 75.1 & 62.9 & 50.7 & 60.0 & 53.4 & 44.5 & 61.3\\
& STAF~\cite{raaj2019efficient}                                           	           & - & - & - & - & - & - & - & 62.7 \\ 
& STEmbedding~\cite{jin2019multi}                                           	           & 78.7 & 79.2 & 71.2 & 61.1 & \bf 74.5 & \bf 69.7 & \bf 64.5 & \bf 71.8 \\\cline{2-10}
\hline
\multirow{5}{*}{\rotatebox{90}{Top-down}} 
& Detect\&Track~\cite{girdhar2018detecttrack}           & 61.7 & 65.5 & 57.3 & 45.7 & 54.3 & 53.1 & 45.7 & 55.2\\ 
& PoseFlow~\cite{xiu2018poseflow}          						  & 59.8 & 67.0 & 59.8 & 51.6 & 60.0 & 58.4 & 50.5 & 58.3\\
& FastPose~\cite{zhang2019fastpose}                                           	           & - & - & - & - & - & - & - & 63.2\\
& FlowTrack~\cite{xiao2018simple}                                           	           & 73.9 & 75.9 & 63.7 & 56.1 & 65.5 & 65.1 & 53.5 & 65.4\\ 
& \cellcolor[gray]{0.9}Our approach                      & \cellcolor[gray]{0.9}\bf 80.5 & \cellcolor[gray]{0.9}\bf 80.9 & \cellcolor[gray]{0.9}\bf 71.6 & \cellcolor[gray]{0.9}\bf 63.8 & \cellcolor[gray]{0.9}70.1 & \cellcolor[gray]{0.9}68.2 & \cellcolor[gray]{0.9}62.0 & \cellcolor[gray]{0.9}\bf 71.6 \\
\hline
\end{tabular}}
\end{center}
\vspace{-7mm}
\caption{\small \it Joint tracking ({\bf MOTA}) on {\bf PoseTrack2017 val}. \vspace{-2mm}}
\label{tab:sota_mota_posetrack2017}
\end{table}


\begin{table}
\begin{center}
\resizebox{\columnwidth}{!}{
\begin{tabular}{|c|l|c|c|c|c|c|c|c|c|}
\cline{2-10}
\multicolumn{1}{c|}{} & Method & Head & Sho & Elb & Wri & Hip & Kne & Ank & \bf Avg \\
\hline
\multirow{2}{*}{\rotatebox{90}{B-U}} 
& STAF~\cite{raaj2019efficient}									& - & - & - & 64.7 & - & - & 62.0 & 70.4 \\
& TML++~\cite{hwang2019pose}                                          	           & - & - & - & - & - & - & - & 74.6\\ \cline{2-10}
\hline
\multirow{2}{*}{\rotatebox{90}{T-D}}  & PT\_CPN++~\cite{yu2018multi}           & 82.4 & \bf 88.8 & 86.2 & \bf 79.4 & 72.0 & \bf 80.6 & \bf 76.2 & 80.9 \\
& \cellcolor[gray]{0.9}Our approach                      & \cellcolor[gray]{0.9} \bf 84.9 & \cellcolor[gray]{0.9} 87.4 & \cellcolor[gray]{0.9}\bf 84.8 & \cellcolor[gray]{0.9}\bf 79.2 & \cellcolor[gray]{0.9}\bf 77.6 & \cellcolor[gray]{0.9} 79.7 & \cellcolor[gray]{0.9} 75.3 & \cellcolor[gray]{0.9}\bf 81.5\\
\hline
\end{tabular}}
\end{center}
\vspace{-7mm}
\caption{\small \it Joint detection ({\bf AP}) on {\bf PoseTrack2018 val}. \vspace{-2mm}}
\label{tab:sota_map_posetrack2018}
\end{table}


\begin{table}
\begin{center}
\resizebox{\columnwidth}{!}{
\begin{tabular}{|c|l|c|c|c|c|c|c|c|c|}
\cline{2-10}
\multicolumn{1}{c|}{} & Method & Head & Sho & Elb & Wri & Hip & Kne & Ank & \bf Avg \\
\hline
\multirow{2}{*}{\rotatebox{90}{B-U}} 
& STAF~\cite{raaj2019efficient}                                           	           & - & - & - & - & - & - & - & 60.9 \\
& TML++~\cite{hwang2019pose}                                          	           & \bf 76.0 & \bf 76.9 & 66.1 & 56.4 & \bf 65.1 & 61.6 & 52.4 & 65.7\\ \cline{2-10}
\hline
\multirow{2}{*}{\rotatebox{90}{T-D}}  & PT\_CPN++~\cite{yu2018multi}           & 68.8 & 73.5 & 65.6 & 61.2 & 54.9 & 64.6 & 56.7 & 64.0 \\ \cline{2-10}
& \cellcolor[gray]{0.9}Our approach                      & \cellcolor[gray]{0.9} 74.2 & \cellcolor[gray]{0.9}76.4 & \cellcolor[gray]{0.9}\bf 71.2 & \cellcolor[gray]{0.9}\bf 64.1 & \cellcolor[gray]{0.9} 64.5 & \cellcolor[gray]{0.9}\bf 65.8 & \cellcolor[gray]{0.9}\bf 61.9 & \cellcolor[gray]{0.9}\bf 68.7 \\
\hline
\end{tabular}}
\end{center}
\vspace{-7mm}
\caption{\small \it Joint tracking ({\bf MOTA}) on {\bf PoseTrack2018 val}. \vspace{-2mm}}
\label{tab:sota_mota_posetrack2018}
\end{table}

\begin{table}
\begin{center}
\resizebox{\columnwidth}{!}{
\begin{tabular}{|l|l|c|c|c|c|}
\hline
Method & Additional Data & wrists AP & ankles AP & \bf Total AP & \bf Total MOTA \\
\hline
JointFlow~\cite{doering2018joint} & COCO & 53.1 & 50.4 & 63.4 & 53.1\\
TML++~\cite{hwang2019pose} & COCO & 60.9 & 56.0 & 67.8 & 54.5\\
FlowTrack~\cite{xiao2018simple}  & COCO & 71.5 & 65.7 & 74.6 & 57.8\\
HRNet~\cite{sun2019deep} & COCO & \bf 72.0 & \bf 67.0 & \bf 75.0 & 57.9\\
POINet~\cite{ruan2019poinet} & COCO & 69.5 & 67.2 & 72.5 & 58.4\\ 
KeyTrack~\cite{snower201915} & COCO & 71.9 & 65.0 & 74.0 & 61.2\\
\hline
\cellcolor[gray]{0.9}Our approach  & \cellcolor[gray]{0.9}COCO & \cellcolor[gray]{0.9} 69.8 & \cellcolor[gray]{0.9} 65.9 & \cellcolor[gray]{0.9} \bf 74.1 & \cellcolor[gray]{0.9} \bf 64.1\\
\hline
\end{tabular}}
\end{center}
\vspace{-7mm}
\caption{\small \it Results from the {\bf PoseTrack2017 test} leaderboard~\cite{posetrack2017}. \vspace{-2mm}}
\label{tab:sota_posetrack2017_test}
\end{table}


\begin{table}
\begin{center}
\resizebox{\columnwidth}{!}{
\begin{tabular}{|l|l|c|c|c|c|}
\hline
Method & Additional Data & wrists AP & ankles AP & \bf Total AP & \bf Total MOTA \\
\hline
TML++~\cite{hwang2019pose}  & COCO & 60.2 & 56.8 & 67.8 & 54.9\\
PT\_CPN++~\cite{yu2018multi} & COCO + Other & 68.2 & 66.1 & 70.9 & 57.4\\
FlowTrack~\cite{xiao2018simple} & COCO + Other & \bf 73.0 & \bf 69.0 &\bf  74.0 & 61.4\\
\hline
\cellcolor[gray]{0.9}Our approach  & \cellcolor[gray]{0.9}COCO & \cellcolor[gray]{0.9} 69.8 & \cellcolor[gray]{0.9} 67.1 & \cellcolor[gray]{0.9} \bf 73.5 & \cellcolor[gray]{0.9} \bf 64.3\\
\hline
\end{tabular}}
\end{center}
\vspace{-7mm}
\caption{\small \it Results from the {\bf PoseTrack2018 test} leaderboard~\cite{posetrack2018}. \vspace{-2mm}}
\label{tab:sota_posetrack2018_test}
\end{table}


\begin{figure*}
\centering
    \includegraphics[width=\linewidth]{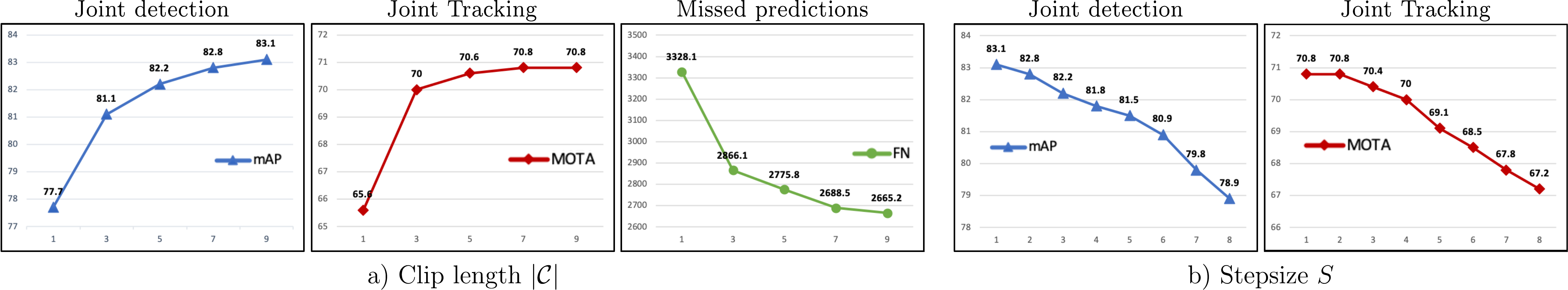}
    \vspace{-7mm}
    \caption{\small \it Results for different values of clips lengths $|\mathcal{C}|$ (a) and stepsize $S$. \vspace{-3mm}} 
    \label{fig:windowsize_stepsize}
\end{figure*}

\subsection{Analysis of our approach} \label{sec:analysis}
We now analyze our approach and our hyper-parameter choices. For simplicity, we run our experiments only on the validation set of PoseTrack2017, using the settings described in sec.~\ref{sec:impl_details}. Unless specified, we do not employ our spatial-temporal merging procedure (sec.~\ref{sec:refinement}) to keep our analysis transparent, as this corrects some mistakes.

\newcommand{\cmark}{\ding{51}}
\begin{table}
	\centering
	\resizebox{\columnwidth}{!}{
	\begin{tabular}{| c | c  c  c | c : c |}
		\hline
		\multirow{3}{*}{\makecell{Backbone:\\HRNet}} & \multirow{3}{*}{Linking} &  Spatial & Temporal & \multirow{3}{*}{\makecell{Detection \\ mAP}} & \multirow{3}{*}{\makecell{Tracking \\ MOTA}} \\
		 & & Merge & Merge & & \\
		  &  & (sec.~\ref{sec:refinement}) & (sec.~\ref{sec:refinement}) & & \\ \hline
		Base 2D & oks-gbm & & & 77.7 & 65.6\\ 
		\hdashline
		 \multirow{5}{*}{\makecell{Our 3D \\ (sec.~\ref{sec:3DHRnet})}} & none & & & 82.3 & -\\
		 & sec.~\ref{sec:tracking} & & & 83.1 & 70.8 \\
		 & sec.~\ref{sec:tracking} & \cmark & & 83.5 & 71.4 \\
		 & sec.~\ref{sec:tracking} &  &\cmark & 83.4 & 71.1 \\
		 & sec.~\ref{sec:tracking} & \cmark & \cmark & \bf 83.8 & \bf 71.6 \\
		\hline
	\end{tabular}}
	\vspace{-3mm}
	\caption{\small \it Ablation study on the components of our approach. In line 3, we test our Video Tracking pipeline paired with a Baseline merge that always selects the hypothesis with the highest score.\vspace{-4mm}}
	\label{tab:ablation_study}
\end{table}

\paragraph{Ablation study.} Here we evaluate the different components of our approach and quantify how much each of them contributes to the model's final performance (table~\ref{tab:ablation_study}). First, we compare against a baseline 2D HRNet model~\cite{sun2019deep} that runs on each frame independently. This baseline model achieves a mAP of 77.7; this is substantially lower compared to our most basic 3D HRNet (82.3 mAP), which does not perform any tracking and just uses OKS-based NMS over the hypotheses. This big improvement is due to our model being able to predict joints in frames where the person detector failed to localized the person. 

When our 3D HRNet is paired with our video tracking pipeline (sec.~\ref{sec:tracking}) and the baseline merge, it improves MOTA considerably compared to the same 2D HRNet baseline paired with the popular OKS-based greedy bipartite matching ({\it oks-gbm}) algorithm that links pose predictions over time~\cite{girdhar2018detecttrack,xiao2018simple}. Interestingly, this also improves mAP over our 3D HRNet with no tracking (+0.8 mAP).
Finally, when we substitute the baseline merge with our procedure (sec.~\ref{sec:refinement}), the results further improve: both spatial and temporal merges are beneficial and complementary, bringing our full model performance to 83.8 mAP and 71.6 MOTA, almost a 10\% improvement over the strong baseline.

 \vspace{-3mm}
\paragraph{Clip length $|\mathcal{C}|$.}
Our 3D HRNet operates on spatial-temporal tubes of length $|\mathcal{C}|$. In sec.~\ref{sec:impl_details}, we set this value to 9, so that both our Clip Tracking Network and our Video Tracking pipeline can greatly benefit from rich temporal information. Here we examine how performance changes as we change this hyperparameter (fig.~\ref{fig:windowsize_stepsize}{\color{red}a}). Setting $|\mathcal{C}|=1$ is equivalent to running the baseline 2D HRNet presented in the previous section and it achieves the lowest performance among all variations.
Interestingly, the largest improvement is brought by moving from $1$ to $3$, which indicates that little temporal information is already sufficient to compensate for many failures of the person detector. 
Further increasing $|\mathcal{C}|$ leads to a slow, but steady improvement in both mAP and MOTA, as the model can recover from even more mistakes. We quantitatively show this recovery in fig.~\ref{fig:windowsize_stepsize}{\color{red}a}, where the number of false negatives decreases as $|\mathcal{C}|$ increases. 


\begin{figure*}
    \includegraphics[width=1.0\textwidth]{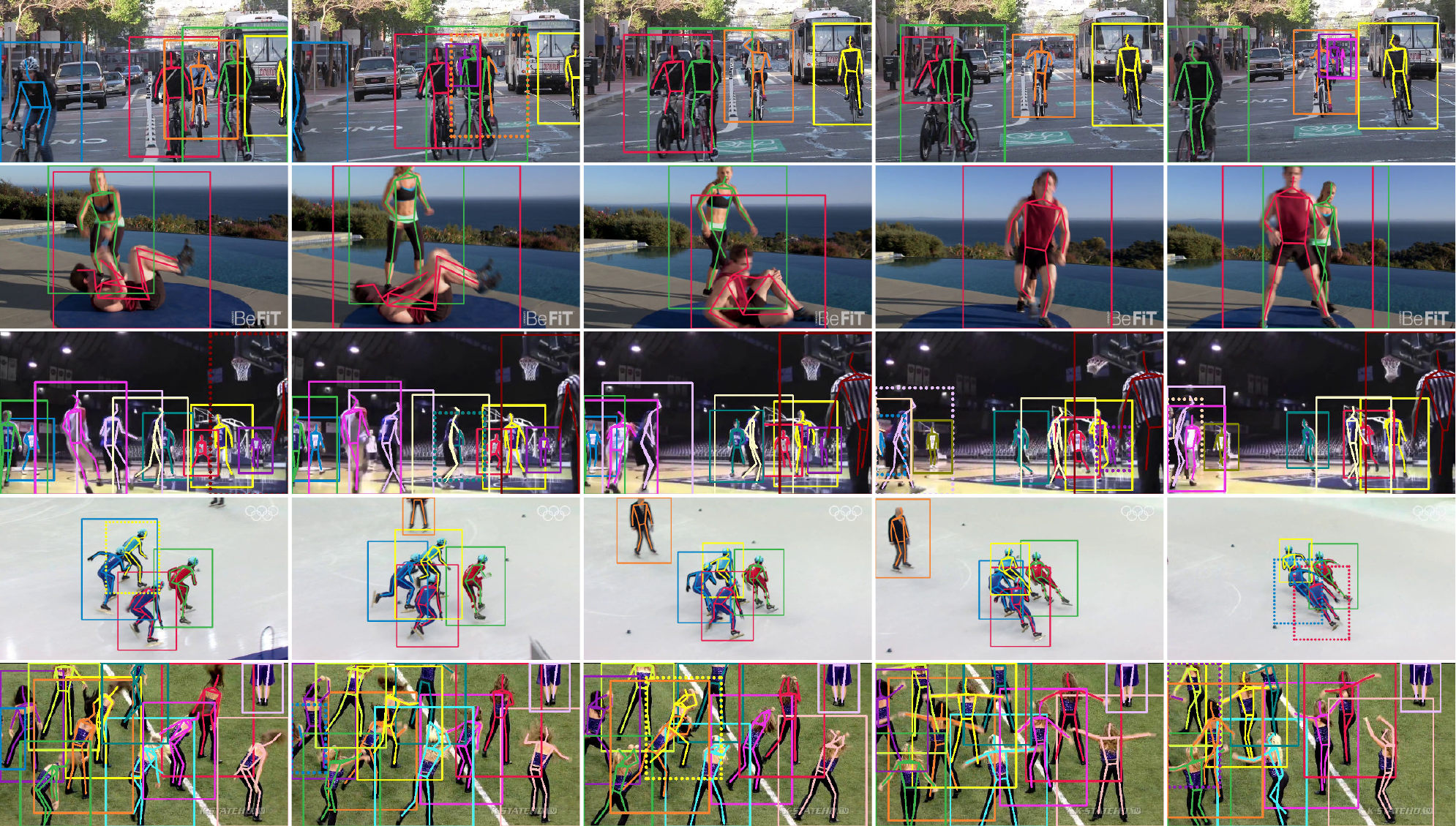}
    \vspace{-4mm}
    \caption{\small \it Visualization of the output of our approach on five videos from the PoseTrack dataset. Bounding boxes and poses are color coded using the track id predicted by our model. Solid bounding boxes indicate that the instance was localized by the person detector, while dotted bounding boxes were originally missed by the detector, but recovered by our approach. \vspace{-4mm}}
    \label{fig:video_results}
\end{figure*}

\begin{table}
	\centering
	\resizebox{0.8\columnwidth}{!}{
	\begin{tabular}{| l | c : ccc|}
		\hline
		
		HRNet: 3D filters & None & Early (ours) & Last & All \\ \hline
		mAP & 77.7 & \bf 81.1 & 80.6 & 79.3 \\
		MOTA & 65.6 & \bf 70.0 & 69.2 & 68.0\\\hline
	\end{tabular}}
	\vspace{-2mm}
	\caption{\small \it Results from different HRNet architecture as Clip Tracking Network, which differ in where they have 3D temporal filters. \vspace{-7mm}}
	\label{tab:different_network}
\end{table}

 \vspace{-3mm}
\paragraph{Step size $S$.}
In sec.~\ref{sec:impl_details}, we set this to 1, so that our approach can use every frame of a video as keyframe and collect the largest set of pose hypotheses. This procedure, however, may be too expensive for some applications and here we evaluate how the performance changes as we improve the runtime by reducing the number of keyframes (i.e., increase the stepsize). Increasing the value of $S$ leads to a linear speed up by a factor $S$, as the two most expensive components of our approach (person detector and 3D HRnet) now run only every $S$ frames. 
As expected, results (fig.~\ref{fig:windowsize_stepsize}{\color{red}b}) for both joint detection and tracking decrease as we increase $S$, as the model looses its temporal benefits. 
Nevertheless, they decreases slowly and even when we run our fastest inference with the largest step size, the model still achieves competitive performance (mAP 78.9 and MOTA 67.2), on par with that of many state-of-the-art models (table~\ref{tab:sota_map_posetrack2017}). 
Furthermore, note out how these results are better than those of our baseline 2D HRNet (mAP 77.7 and MOTA 65.6, fig.~\ref{fig:windowsize_stepsize}{\color{red}a}, $|\mathcal{C}|=1$), yet this 3D model is effectively faster, as it runs its person detector only once every 8 frames, as opposed to all frames, as done by 2D HRNet.  

 \vspace{-3mm}
\paragraph{Network design.}
Our 3D HRNet architecture uses 3D convolutions in its early 2 stages (sec.~\ref{sec:3DHRnet}), as these are the best suited to learn the low-level correspondence needed to correctly link the joints of the same person within a tube. 
In this section we evaluate different network designs: our design (Early), a 3D HRNet architecture with 3D filters in its last stage (Last), which learn to smooth joint predictions over small temporal windows, and a fully 3D HRNet architecture (All), that balances learning good temporal correspondences and spatially smooth joint predictions.
As training a full 3D HRNet requires a considerable amount of GPU memory, we experiment here with a lightweight setup with $|\mathcal{C}|=3$. 
Results are presented in table~\ref{tab:different_network}. For reference, we report the mAP performance of a standard 2D HRnet without any 3D filter. Adding 3D filters, no matter the location, always improves over the simple 2D architecture. Among the different choices, ``Early'' achieves the best performance for both detection and tracking, validating our design.

 \vspace{-3mm}
\paragraph{Dependency on person detector.}
Like all top-down methods, our approach is also limited by the accuracy of the employed person detector. However, we believe that our approach is significantly less sensitive than others in the literature, as it can recover missed predictions using its temporal reasoning. To validate this, we evaluate how well the propagation of detection boxes to neighboring frames allows the model to improve recall. We experiment on the validation set of PoseTrack2018, as the 2017 set does not have bounding box annotations. We compare our 3D approach against its 2D counterpart, using two different backbones (table~\ref{tab:det}). Results show that: (i) our 3D approach can indeed recover a substantial number of missed predictions (+4-7\% recall) and (ii) it can even raise the recall of a weaker detector (3D MobileNet-V2, recall 83) on par with that of a much stronger model (2D ResNet-101, recall 82.9).

\begin{table}
\centering 
\resizebox{0.7\columnwidth}{!}{
\begin{tabular}{| c | c| c| }
\hline
Person detector & Base 2D & Our 3D \\\hline
{\it Strong} ResNet-101 & 82.9 & 86.5 \\
{\it Weaker} MobileNet-V2 & 77.6 & 83.0 \\\hline
\end{tabular}}
\vspace{-3mm}
 \caption{\small \it Person bounding box recall on PoseTrack 2018. \vspace{-3mm}}
 \label{tab:det}
\end{table}

\section {Conclusion} \label{sec:concl} 
We have presented a novel top-down approach for multi-person pose estimation and tracking in videos. Our approach can recover from failures of its person detector by propagating known person locations through time and by searching for poses in them. Our approach consists of three components. Clip Tracking Network was used to jointly perform joint pose estimation and tracking on small video clips. Then, Video Tracking Pipeline was used to merge tracklets predicted by Clip Tracking Network, when these belonged to the same person. Finally, Spatial-Temporal Merging was used to refine the joint locations based on a spatial-temporal consensus procedure over multiple detections for the same person. We showed that this approach is capable of correctly predicting people poses, even on very hard scenes containing severe occlusion and entanglements (fig.~\ref{fig:video_results}). Finally, we showed the straight of our approach by achieving state-of-the-art results on both joint detection and tracking, on both the PoseTrack 2017 and 2018 datasets, and against all top-down and bottom-down approaches. 


{\small
\bibliographystyle{ieee_fullname}
\bibliography{egbib}

\begin{thebibliography}{10}\itemsep=-1pt

\bibitem{posetrack2017}
Posetrack 2017: Leader board.
\newblock \url{https://posetrack.net/leaderboard.php}, 2017.

\bibitem{posetrack2018}
Posetrack 2018: Leader board.
\newblock
  \url{https://posetrack.net/workshops/eccv2018/posetrack_eccv_2018_results.html},
  2018.

\bibitem{PoseTrack}
Mykhaylo Andriluka, Umar Iqbal, Eldar Insafutdinov, Leonid Pishchulin, Anton
  Milan, Juergen Gall, and Bernt Schiele.
\newblock Pose{T}rack: {A} benchmark for human pose estimation and tracking.
\newblock In {\em CVPR}, 2018.

\bibitem{andriluka14cvpr}
Mykhaylo Andriluka, Leonid Pishchulin, Peter Gehler, and Bernt Schiele.
\newblock 2d human pose estimation: New benchmark and state of the art
  analysis.
\newblock In {\em CVPR}, 2014.

\bibitem{cao2018openpose}
Zhe Cao, Tomas Simon, Shih-En Wei, and Yaser Sheikh.
\newblock Realtime multi-person 2d pose estimation using part affinity fields.
\newblock In {\em CVPR}, 2017.

\bibitem{carreira2017quo}
Joao Carreira and Andrew Zisserman.
\newblock Quo vadis, action recognition? a new model and the kinetics dataset.
\newblock In {\em CVPR}, 2017.

\bibitem{chen2018cascaded}
Yilun Chen, Zhicheng Wang, Yuxiang Peng, Zhiqiang Zhang, Gang Yu, and Jian Sun.
\newblock Cascaded pyramid network for multi-person pose estimation.
\newblock In {\em CVPR}, 2018.

\bibitem{comaniciu2002mean}
Dorin Comaniciu and Peter Meer.
\newblock Mean shift: A robust approach toward feature space analysis.
\newblock {\em TPAMI}, 5:603--619, 2002.

\bibitem{dai17iccv}
Jifeng Dai, Haozhi Qi, Yuwen Xiong, Yi Li, Guodong Zhang, Han Hu, and Yichen
  Wei.
\newblock Deformable convolutional networks.
\newblock In {\em ICCV}, 2017.

\bibitem{dijkstra1959note}
Edsger~W Dijkstra.
\newblock A note on two problems in connexion with graphs.
\newblock {\em Numerische mathematik}, 1(1):269--271, 1959.

\bibitem{doering2018joint}
Andreas Doering, Umar Iqbal, and Juergen Gall.
\newblock Joint flow: Temporal flow fields for multi person tracking.
\newblock In {\em BMVC}, 2018.

\bibitem{feichtenhofer2016spatiotemporal}
Christoph Feichtenhofer, Axel Pinz, and Richard Wildes.
\newblock Spatiotemporal residual networks for video action recognition.
\newblock In {\em NIPS}, 2016.

\bibitem{girdhar2018detecttrack}
Rohit Girdhar, Georgia Gkioxari, Lorenzo Torresani, Manohar Paluri, and Du
  Tran.
\newblock {Detect-and-Track: Efficient Pose Estimation in Videos}.
\newblock In {\em CVPR}, 2018.

\bibitem{he2017mask}
Kaiming He, Georgia Gkioxari, Piotr Doll{\'a}r, and Ross Girshick.
\newblock Mask r-cnn.
\newblock In {\em ICCV}, 2017.

\bibitem{hwang2019pose}
Jihye Hwang, Jieun Lee, Sungheon Park, and Nojun Kwak.
\newblock Pose estimator and tracker using temporal flow maps for limbs.
\newblock {\em IJCNN}, pages 1--8, 2019.

\bibitem{insafutdinov16eccv}
Eldar Insafutdinov, Leonid Pishchulin, Bjoern Andres, Mykhaylo Andriluka, and
  Bernt Schiele.
\newblock Deepercut: A deeper, stronger, and faster multi-person pose
  estimation model.
\newblock In {\em ECCV}, 2016.

\bibitem{jin2019multi}
Sheng Jin, Wentao Liu, Wanli Ouyang, and Chen Qian.
\newblock Multi-person articulated tracking with spatial and temporal
  embeddings.
\newblock In {\em CVPR}, 2019.

\bibitem{kuhn1955hungarian}
Harold~W Kuhn.
\newblock The hungarian method for the assignment problem.
\newblock {\em Naval research logistics quarterly}, 2(1-2):83--97, 1955.

\bibitem{lin2014microsoft}
Tsung-Yi Lin, Michael Maire, Serge Belongie, James Hays, Pietro Perona, Deva
  Ramanan, Piotr Doll{\'a}r, and C~Lawrence Zitnick.
\newblock Microsoft coco: Common objects in context.
\newblock In {\em ECCV}, 2014.

\bibitem{milan2016mot16}
Anton Milan, Laura Leal-Taix{\'e}, Ian Reid, Stefan Roth, and Konrad Schindler.
\newblock M{OT}16: A benchmark for multi-object tracking.
\newblock {\em arXiv preprint arXiv:1603.00831}, 2016.

\bibitem{newell2017associative}
Alejandro Newell, Zhiao Huang, and Jia Deng.
\newblock Associative embedding: End-to-end learning for joint detection and
  grouping.
\newblock In {\em NIPS}, 2017.

\bibitem{newell16eccv}
Alejandro Newell, Kaiyu Yang, and Jia Deng.
\newblock Stacked hourglass networks for human pose estimation.
\newblock In {\em ECCV}, 2016.

\bibitem{papandreou17cvpr}
George Papandreou, Tyler Zhu, Nori Kanazawa, Alexander Toshev, Jonathan
  Tompson, Chris Bregler, and Kevin Murphy.
\newblock Towards accurate multi-person pose estimation in the wild.
\newblock In {\em CVPR}, 2017.

\bibitem{pishchulin16cvpr}
Leonid Pishchulin, Eldar Insafutdinov, Siyu Tang, Bjoern Andres, Mykhaylo
  Andriluka, Peter~V Gehler, and Bernt Schiele.
\newblock Deepcut: Joint subset partition and labeling for multi person pose
  estimation.
\newblock In {\em CVPR}, 2016.

\bibitem{raaj2019efficient}
Yaadhav Raaj, Haroon Idrees, Gines Hidalgo, and Yaser Sheikh.
\newblock Efficient online multi-person 2d pose tracking with recurrent
  spatio-temporal affinity fields.
\newblock In {\em CVPR}, 2019.

\bibitem{ruan2019poinet}
Weijian Ruan, Wu Liu, Qian Bao, Jun Chen, Yuhao Cheng, and Tao Mei.
\newblock Poinet: pose-guided ovonic insight network for multi-person pose
  tracking.
\newblock In {\em ACM Multimedia}, 2019.

\bibitem{ronchi17iccv}
Matteo Ruggero~Ronchi and Pietro Perona.
\newblock Benchmarking and error diagnosis in multi-instance pose estimation.
\newblock In {\em ICCV}, 2017.

\bibitem{singh18nips}
Bharat Singh, Mahyar Najibi, and Larry~S. Davis.
\newblock S{NIPER}: Efficient multi-scale training.
\newblock In {\em NIPS}, 2018.

\bibitem{snower201915}
Michael Snower, Asim Kadav, Farley Lai, and Hans~Peter Graf.
\newblock 15 keypoints is all you need.
\newblock {\em arXiv preprint arXiv:1912.02323}, 2019.

\bibitem{sun2019deep}
Ke Sun, Bin Xiao, Dong Liu, and Jingdong Wang.
\newblock Deep high-resolution representation learning for human pose
  estimation.
\newblock In {\em CVPR}, 2019.

\bibitem{sun2019arxiv}
Ke Sun, Yang Zhao, Borui Jiang, Tianheng Cheng, Bin Xiao, Dong Liu, Yadong Mu,
  Xinggang Wang, Wenyu Liu, and Jingdong Wang.
\newblock High-resolution representations for labeling pixels and regions.
\newblock {\em arXiv preprint arXiv:1904.04514}, 2019.

\bibitem{wang19arxiv}
Jingdong Wang, Ke Sun, Tianheng Cheng, Borui Jiang, Chaorui Deng, Yang Zhao,
  Dong Liu, Yadong Mu, Mingkui Tan, Xinggang Wang, Wenyu Liu, and Bin Xiao.
\newblock Deep high-resolution representation learning for visual recognition.
\newblock {\em arXiv preprint arXiv:1908.07919}, 2019.

\bibitem{wei16cvpr}
Shih-En Wei, Varun Ramakrishna, Takeo Kanade, and Yaser Sheikh.
\newblock Convolutional pose machines.
\newblock In {\em CVPR}, 2016.

\bibitem{xiao2018simple}
Bin Xiao, Haiping Wu, and Yichen Wei.
\newblock Simple baselines for human pose estimation and tracking.
\newblock In {\em ECCV}, 2018.

\bibitem{xiu2018poseflow}
Yuliang Xiu, Jiefeng Li, Haoyu Wang, Yinghong Fang, and Cewu Lu.
\newblock {Pose Flow}: Efficient online pose tracking.
\newblock In {\em BMVC}, 2018.

\bibitem{yu2018multi}
Dongdong Yu, Kai Su, Jia Sun, and Changhu Wang.
\newblock Multi-person pose estimation for pose tracking with enhanced cascaded
  pyramid network.
\newblock In {\em ECCVW}, 2018.

\bibitem{zhang2019fastpose}
Jiabin Zhang, Zheng Zhu, Wei Zou, Peng Li, Yanwei Li, Hu Su, and Guan Huang.
\newblock Fastpose: Towards real-time pose estimation and tracking via
  scale-normalized multi-task networks.
\newblock {\em arXiv preprint arXiv:1908.05593}, 2019.

\end{thebibliography}
}

\end{document}